\documentclass[10pt,twocolumn,letterpaper]{article}

\def\smodel{PAL\xspace}
\def\lmodel{Pseudo Action Localization\xspace}

\usepackage{cvpr}              % To produce the CAMERA-READY version
\usepackage{graphicx}
\usepackage{amsmath}
\usepackage{amssymb}
\usepackage{booktabs}
\usepackage{caption}
\usepackage{subcaption}
\usepackage[table,dvipsnames]{xcolor}
\usepackage{bm}
\usepackage{mathtools}
\usepackage{multirow}

\usepackage{arydshln}
\usepackage{verbatim}
\usepackage{pifont}
\newcommand{\xmark}{\ding{55}}%
\usepackage{soul}
\usepackage{lipsum}
\usepackage[accsupp]{axessibility}

\newcommand\mypull{\stackrel{\mathclap{\normalfont\mbox{\footnotesize pull}}}{\rightarrow \leftarrow}}

\newcommand{\CUT}[1]{}
\newcommand\inv[1]{#1\raisebox{0.9ex}{$\scriptscriptstyle\text{-}\,\!1$}}

\newcommand*{\belowrulesepcolor}[1]{%
  \noalign{%
    \kern-\belowrulesep
    \begingroup
      \color{#1}%
      \hrule height\belowrulesep
    \endgroup
  }%
}
\newcommand*{\aboverulesepcolor}[1]{%
  \noalign{%
    \begingroup
      \color{#1}%
      \hrule height\aboverulesep
    \endgroup
    \kern-\aboverulesep
  }%
}

\definecolor{myblue}{RGB}{121,210,251}
\definecolor{myblue2}{RGB}{26,147,252}
\definecolor{mygreen}{RGB}{102,205,127}
\definecolor{myred}{RGB}{255,124,124}
\definecolor{myyellow}{RGB}{249,210,88}

\definecolor{aliceblue}{rgb}{0.94, 0.97, 1.0}
\definecolor{mistyrose}{rgb}{1.0, 0.94, 0.94}
\DeclareRobustCommand{\bluehl}[1]{{\sethlcolor{aliceblue}\hl{#1}}}

\usepackage[pagebackref,breaklinks,colorlinks]{hyperref}
\usepackage{marvosym}

\usepackage[capitalize]{cleveref}
\crefname{section}{Sec.}{Secs.}
\Crefname{section}{Section}{Sections}
\Crefname{table}{Table}{Tables}
\crefname{table}{Tab.}{Tabs.}

\makeatletter
\@namedef{ver@everyshi.sty}{}
\makeatother
\usepackage{pgfplots}
\pgfplotsset{every tick label/.append style={font=\footnotesize}}
\pgfplotsset{compat=1.11,
        /pgfplots/ybar legend/.style={
        /pgfplots/legend image code/.code={%
        \draw[##1,/tikz/.cd,bar width=7pt,yshift=-0.2em,bar shift=0pt]
                plot coordinates {(0cm,0.8em)};},
},
}

\def\cvprPaperID{1024} % *** Enter the CVPR Paper ID here
\def\confName{CVPR}
\def\confYear{2022}

\begin{document}

\title{Unsupervised Pre-training for Temporal Action Localization Tasks}

\author{
Can Zhang$^{1}$\thanks{Work done during an internship at Tencent AI Lab.} \ \ 
Tianyu Yang$^2$ \ 
Junwu Weng$^2$ \ 
Meng Cao$^1$ \ 
Jue Wang$^2$ \ 
Yuexian Zou$^{1\textrm{\Letter}}$\\
$^1$School of Electronic and Computer Engineering, Peking University \quad $^2$Tencent AI Lab \\
{\tt\small zhangcan@pku.edu.cn \quad tianyu-yang@outlook.com \quad WE0001WU@e.ntu.edu.sg}\\
{\tt\small mengcao@pku.edu.cn \qquad arphid@gmail.com \qquad zouyx@pku.edu.cn}
}

\maketitle

%%%%%%%%% ABSTRACT
\begin{abstract}
Unsupervised video representation learning has made remarkable achievements in recent years. However, most existing methods are designed and optimized for video classification. These pre-trained models can be sub-optimal for temporal localization tasks due to the inherent discrepancy between video-level classification and clip-level localization. To bridge this gap, we make the first attempt to propose a self-supervised pretext task, coined as Pseudo Action Localization (PAL) to Unsupervisedly Pre-train feature encoders for Temporal Action Localization tasks (UP-TAL). Specifically, we first randomly select temporal regions, each of which contains multiple clips, from one video as pseudo actions and then paste them onto different temporal positions of the other two videos. The pretext task is to align the features of pasted pseudo action regions from two synthetic videos and maximize the agreement between them. Compared to the existing unsupervised video representation learning approaches, our PAL adapts better to downstream TAL tasks by introducing a temporal equivariant contrastive learning paradigm in a temporally dense and scale-aware manner. Extensive experiments show that PAL can utilize large-scale unlabeled video data to significantly boost the performance of existing TAL methods. Our codes and models will be made publicly available at \url{https://github.com/zhang-can/UP-TAL}.
\end{abstract}

%%%%%%%%% BODY TEXT
\vspace{-0.53cm}
\section{Introduction}
\label{sec:intro}

Model pre-training is an effective technique for training deep networks in many computer vision tasks. The core idea is to learn general representations on large-scale labeled or unlabeled data, and utilize the learned representations to improve the performance of downstream tasks with limited data. This is especially beneficial for tasks that require enormous human effort to annotate data, such as temporal action localization (TAL).

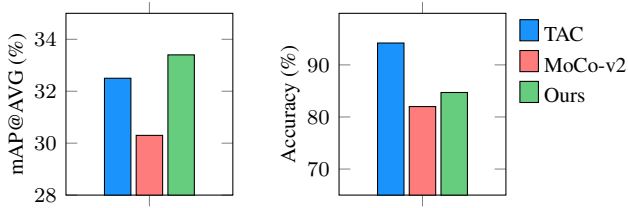
\begin{figure}[t]
\begin{center}
\centering
     \begin{subfigure}[b]{0.20\textwidth}
         \begin{tikzpicture}
			\begin{axis}[
			    ybar,
			    width=3.8cm,
			    height=4.0cm,
			    samples=2,
			    domain=1:2,
			    xtick=data,
			    enlarge x limits={abs=1},
			    xtick={0},
				xticklabels={},
				ymin=28,
				ymax=35,
        ylabel={mAP@AVG (\%)},
        ylabel shift = -4 pt,
        label style={font=\footnotesize}
			]
			\addplot[fill=myblue2] coordinates {
			    (0,32.5) 
			};
			\addplot[fill=myred] coordinates {
			    (0,30.3) 
			};
			\addplot[fill=mygreen] coordinates {
			    (0,33.4) 
			};
			\end{axis}
		 \end{tikzpicture}
         \caption{Action Localization (TAL)}
         \label{fig:teaser_TAL}
     \end{subfigure}
     \hfill
     \begin{subfigure}[b]{0.27\textwidth}
         \centering
         \begin{tikzpicture}
			\begin{axis}[
			    ybar,
			    width=3.8cm,
			    height=4.0cm,
			    samples=2,
			    domain=1:2,
			    xtick=data,
			    ymax=99.9,
			    ymin=65,
			    enlarge x limits={abs=1},
			    xtick={0},
				xticklabels={},
				legend cell align=left,
			    legend pos=outer north east,
			    legend style={font=\footnotesize},
			    legend style={draw=none},
			    ylabel={Accuracy (\%)},
		        ylabel shift = -4 pt,
		        label style={font=\footnotesize}
			]
			\addplot[fill=myblue2] coordinates {
			    (0,94.2) 
			};
			\addplot[fill=myred] coordinates {
			    (0,82.0) 
			};
			\addplot[fill=mygreen] coordinates {
			    (0,84.7) 
			};
			\legend{TAC,MoCo-v2,Ours}
			\end{axis}
		 \end{tikzpicture}
         \caption{Action Classification (TAC)}
         \label{fig:teaser_TAC}
     \end{subfigure}
\end{center}
\vspace{-15pt}
 \caption{\textbf{Comparison of Kinetics-400 pre-trained models by fine-tuning on downstream TAL (ActivityNet v1.3) and TAC (UCF101) datasets.} `TAC' means supervised TAC pre-training, and we treat MoCo-v2~\cite{chen2020improved} with video input as our baseline. Instance-level discrimination is not well-aligned with TAL, thus unsupervised pre-training tailored for TAL is on demand.}
\label{fig:intro}
\end{figure}

Despite the prevailing use of ready-made feature extractors \cite{tran2015learning,carreira2017quo,wang2016temporal} pre-trained on temporal action classification (TAC) in TAL, this pre-training strategy is sub-optimal as the {\it inherent discrepancy} between TAC and TAL exists. Without a doubt, this discrepancy impedes further performance improvement of TAL. Though some recent works \cite{Xu_2021_ICCV,alwassel2021tsp,xu2021low} attempt to tackle this issue, they still rely on large-scale annotated video data. Recently, unsupervised pre-training has attracted great attention due to its potentials in exploiting large amounts of unlabeled data. Contrastive learning~\cite{oord2018representation,chen2020simple,he2020momentum,chen2020improved,grill2020bootstrap} is one of the most popular directions that focus on instance discrimination, which pulls instance-level positive pairs closer while repelling negative ones apart in the embedding space. To fill the gap between the upstream pre-training and the downstream tasks, recent contrastive learning methods focus on specifically designing pretext tasks for various downstream image tasks, \textit{e.g.}, object detection~\cite{wang2021dense,yang2021instance,xie2021detco}, semantic segmentation~\cite{wang2021dense,Van_Gansbeke_2021_ICCV}, \textit{etc}. In contrast, the progress of unsupervised pre-training in video domain is relatively lagging behind and most existing methods~\cite{han2020self,alwassel2020self,wang2020self,pan2021videomoco,Jenni_2021_ICCV,qian2021spatiotemporal} are still designed and evaluated for classification tasks.

In this paper, we make the first attempt on unsupervised pre-training for TAL tasks. One possible way~\cite{qian2021spatiotemporal} to achieve this is to directly extend the image contrastive learning idea to the video domain, where a video is treated as an instance and the clips are regarded as views of instances. Those clip embeddings from the same video are pulled closer while those from different videos are pushed apart. Clearly, this way only focuses on instance (video-level) discrimination, \textit{i.e.}, learning \textit{time-invariant} features for specific video instances, which is required by TAC task in essence. In contrast, TAL expects the representations to be \textit{equivariant to temporal translation and scale}. For example, if we change the start time and duration of an action instance in the input video, the output classification responses of TAC should be unchanged, while the output localization predictions of TAL need to be altered accordingly. The inherent discrepancy between these two tasks attracts our attention to question the suitability of the existing instance discrimination paradigm for TAL. Indeed, as shown in Fig.~\ref{fig:intro}, such video-level discrimination is beneficial for TAC tasks, but not well-aligned with TAL tasks. So, it is desirable and challenging to design a new learning scheme that can be transferred well on TAL tasks.

Motivated by the inherent discrepancy between TAC and TAL, we introduce \textit{temporal equivariant} contrastive learning paradigm by designing a new unsupervised pretext task called \textit{\lmodel} (\smodel). Specifically, to mimic the TAL-tailored data with temporal boundaries, we first construct our training set by transforming the existing large-scale TAC datasets in a cheap manner. We randomly crop two temporal regions with random temporal lengths and scales from one video as pseudo actions. Each of these regions includes multiple consecutive clips. Then we paste them onto different temporal positions of other randomly selected background videos. With the preset temporal transformation (paste location, clip length, sampling scale), the model is able to align the pseudo action features of two synthesized videos. Such transformation and alignment process are named as \textit{input-level transformation} and \textit{feature-level equi-transformation} in our paper. 
Moreover, to better align the upstream pre-training pipeline to the downstream TAL architecture, we follow the way of estimating temporal locations in TAL tasks ~\cite{lin2019bmn, lin2018bsn} by applying several layers of temporal convolutions to process the sequential clip-level features. Thereby, the information of surrounding background clips is highly involved in the final output features of pseudo action regions. With the {\it random} paste operation, the diversity of background-involvement is increased. Further, we propose to maximize the agreement between two aligned pseudo action region features such that the learned features are forced to focus on the most discriminative and background-irrelevant parts, thus enhancing their robustness and achieving the equivariance requirement in TAL.

We summarize our main contributions as follows: (1) To our best knowledge, this is the FIRST work focusing on unsupervised pre-training for temporal action localization tasks (UP-TAL). (2) We design an intuitive and effective self-supervised pretext task customized for TAL, called \smodel. A time-equivariant contrastive learning paradigm is also introduced to perform transformed foreground discrimination, customized for TAL representation learning. (3) Extensive experiments on ActivityNet v1.3~\cite{caba2015activitynet}, Charades-STA~\cite{gao2017tall} and THUMOS'14~\cite{idrees2017thumos} datasets show that \smodel transfers well on various downstream TAL-related tasks: Temporal Action Detection (TAD), Action Proposal Generation (APG) and Video Grounding (VG). Notably, our \smodel even surpasses the supervised pre-training when using the same amount of video data.

\section{Related work}
\label{sec:related_work}

\textbf{Contrastive Video Representation Learning.}
Recently, contrastive learning~\cite{oord2018representation,he2020momentum,chen2020simple,chen2020improved,grill2020bootstrap,chen2021exploring,caron2020unsupervised} has gained increasing attention due to its outstanding performance. 
Essentially, these contrast-based methods focus on \textit{instance discrimination}~\cite{wu2018unsupervised}, \textit{i.e.}, distinguishing each instance from the rest. 
Following this direction, recent researches~\cite{wang2020self,pan2021videomoco,qian2021spatiotemporal,yang2020video} extend the contrastive learning idea to the video domain, where clips from the same video are considered as positives and clips from the different videos as negatives. 
Besides, other directions, such as: dense future prediction~\cite{han2020memory,han2019video}, cross-modal supervision~\cite{sun2019videobert,alwassel2020self,han2020self}, \textit{etc}, have also been studied in the literature. Notably, most of these methods are designed for TAC tasks that learn time-invariant features. 
In contrast, we propose a novel pretext task tailored for TAL, which follows a temporal equivariant learning scheme.
A concurrent work~\cite{Jenni_2021_ICCV} also focuses on time-equivariant representation learning. Two clips from different videos but with the same relative transformation (overlap/order) are considered as positive pairs, which promotes detailed learning of motion patterns and thus is beneficial for TAC tasks. Our method differs essentially from it in the fact that positive pairs are constructed from two transformed regions (multiple clips) of the same foreground video but with different backgrounds. This facilitates the learning of TAL-friendly features such that they are robust to background interference but sensitive to temporal transformation (scale and location).

\begin{figure*}[t]
\begin{center}
\begin{subfigure}[b]{0.35\linewidth}
         \centering
\includegraphics[width=\linewidth]{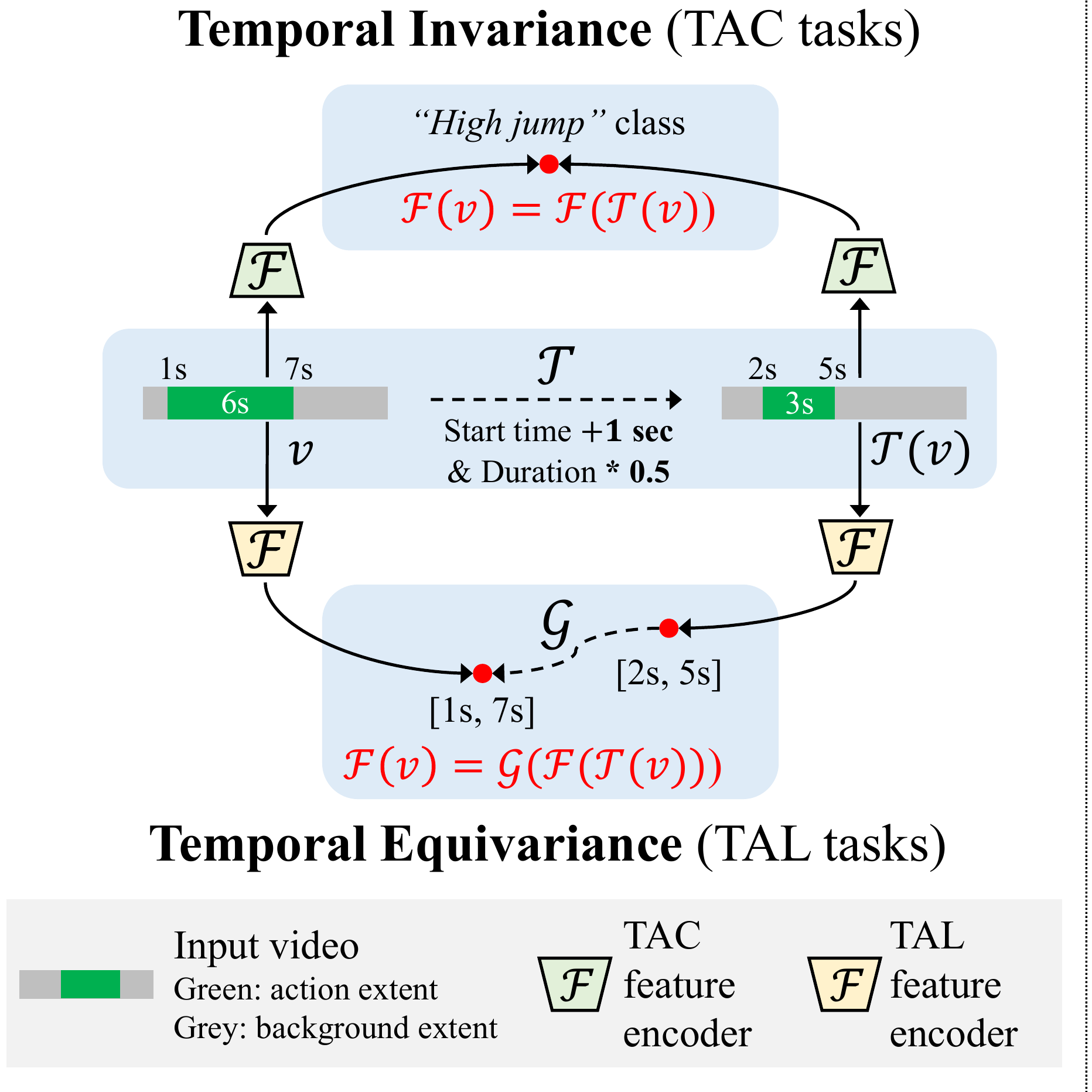}
\caption{}
\label{fig:in-equi}
\end{subfigure}
\begin{subfigure}[b]{0.63\linewidth}
         \centering
\includegraphics[width=\linewidth]{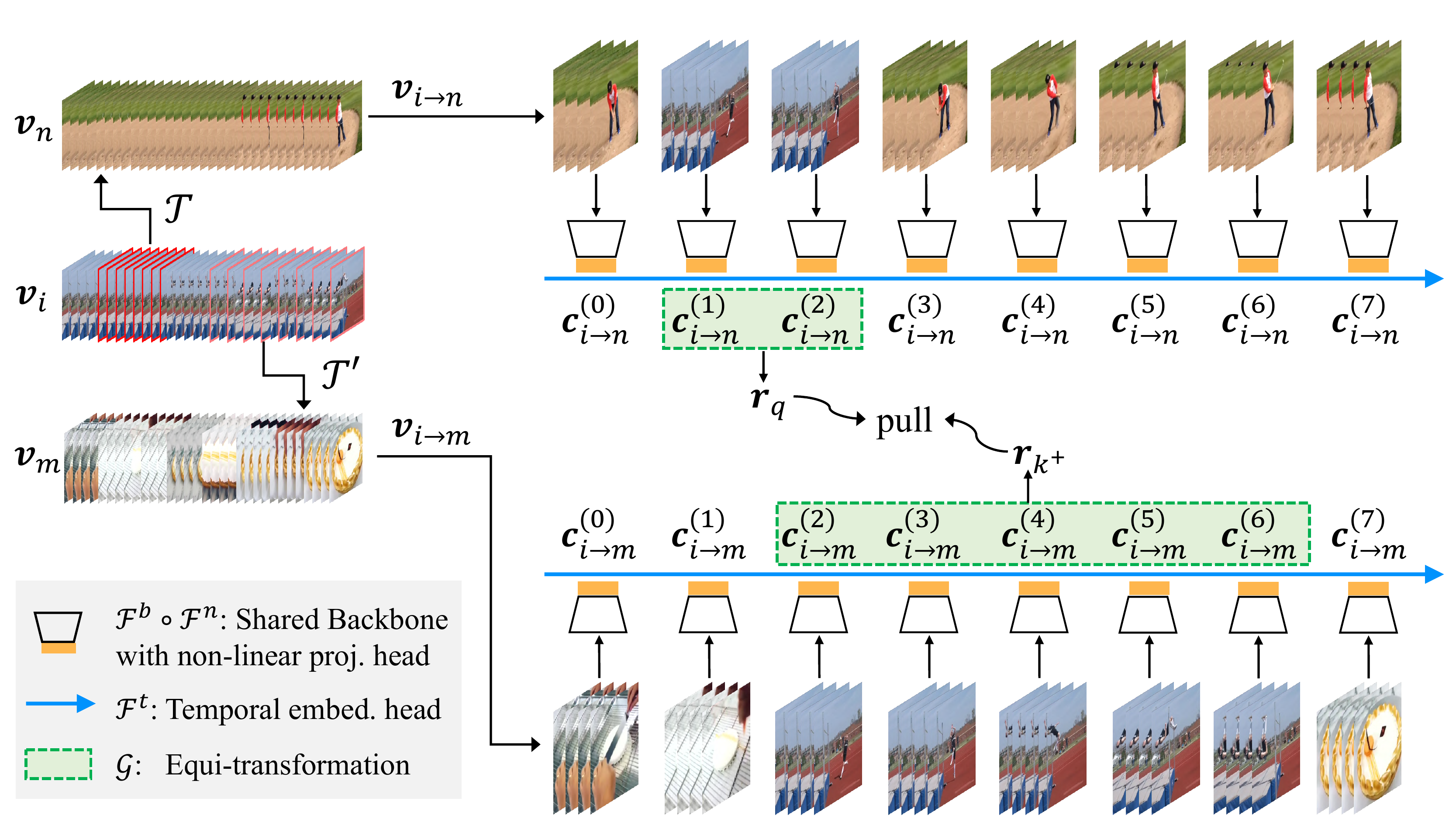}
\caption{}
\label{fig:PAL}
\end{subfigure}
\end{center}
\vspace{-0.5cm}
 \caption{\textbf{(a) Schematic depiction of temporal invariance \textit{vs.} temporal equivariance. (b) Overview of our \smodel pretext task.} Given a video $\bm{v}_i$, we randomly sample two pseudo action regions from it and then paste them onto another two pseudo background videos at various temporal locations and scales. \smodel learns temporal equivariant features by aligning pseudo action region features and maximizing the agreement between region features of the same video but with different backgrounds. Negatives are omitted for brevity.}
\label{fig:UP-TAL}
\end{figure*}

\textbf{Temporal Action Localization (TAL) Tasks.} Unlike TAC~\cite{wang2016temporal,carreira2017quo,tu2018semantic,tu2019action,zhang2019pan,lin2019tsm,feichtenhofer2019slowfast}, the target of TAL is to temporally localize the action of interest in untrimmed videos. In general, TAL covers a range of tasks, such as: \textit{Action Proposal Generation} (APG), \textit{Temporal Action Detection} (TAD) and \textit{Video Grounding} (VG), \textit{etc}. APG aims at generating temporal proposals which are likely to contain human actions. Previous methods design temporal anchor instances for feature sequences~\cite{heilbron2016fast,lin2017temporal,buch2017sst} or directly predict boundary probabilities~\cite{lin2018bsn,lin2019bmn}. TAD aims at predicting the temporal extent as well as the class labels of action instances. Most existing fully-supervised TAD methods ~\cite{xu2017r,chao2018rethinking,long2019gaussian,lin2019bmn,xu2020g,bai2020boundary,chen2020afnet} integrate the proposal generation and classification procedures in a unified network. Some recent works have also designed TAD algorithms under weaker supervision~\cite{wang2017untrimmednets,nguyen2018weakly,zeng2019breaking,zhang2021cola,zhong2018step}. VG, \textit{a.k.a}, text-to-video temporal grounding, aims to localize the time interval corresponding to a given text query. The current literature can be roughly divided into two categories, namely proposal-based~\cite{chen2020hierarchical,ghosh-etal-2019-excl,yuan2019find,zhang2020span} and proposal-free~\cite{anne2017localizing,liu2018cross,mun2020local,cao2021pursuit} architectures. For these TAL tasks, we choose three representative works (BMN~\cite{lin2019bmn}, G-TAD~\cite{xu2020g} and LGI~\cite{mun2020local}) with officially released code to validate the efficacy of our PAL.

\textbf{Supervised Pre-training for TAL.} 
Due to the GPU memory constraint, the common practice in TAL is to first pre-train a feature encoder on large-scale trimmed TAC datasets, and then use it to extract frame-level or segment-level features in untrimmed TAL videos. 
Inevitably, this will result in a task discrepancy problem, since feature encoders are trained on TAC while used for TAL. This domain gap has not been fully studied though it is common in TAL. Recent advances try to bridge this gap through boundary type classification~\cite{Xu_2021_ICCV}, foreground region classification~\cite{alwassel2021tsp} and end-to-end training~\cite{xu2021low}. Unfortunately, they all belong to the supervised pre-training paradigm and therefore rely on large-scale labeled videos. In contrast, we propose a novel method, for the first time (to our best knowledge), focusing on Unsupervised Pre-training for TAL (UP-TAL). 

\textbf{Cut-Paste for Data Synthesis.} Cut-Paste, which cuts a part of one data sample and pastes it onto another sample, is found to be a useful data augmentation strategy when facing the data shortage issue. It has been widely adopted in supervised learning for object detection~\cite{dvornik2018modeling,dwibedi2017cut}, instance segmentation~\cite{fang2019instaboost,Ghiasi_2021_CVPR}, and self-supervised learning for image/video classification~\cite{zhao2021distilling,Wang_2021_CVPR}, object detection~\cite{yang2021instance} and anomaly detection~\cite{li2021cutpaste}, \textit{etc}. The most recent work relevant to ours is BSP~\cite{Xu_2021_ICCV}, which also synthesizes videos through temporal Cut-Paste. The essential difference lies in the fact that BSP \textit{supervisedly} generates different types of temporal boundaries and learn to predict them to facilitate the learning of video features while our PAL synthesizes videos without using any label information and train the backbone by aligning pseudo action region features from two synthetic videos and maximizing their agreement with temporal equivariant contrastive learning. 

\section{Method}

\subsection{Intuition and Preliminaries}

As mentioned in Sec.~\ref{sec:intro}, the most essential difference between TAC and TAL is that the former requires \textit{temporal invariance} while the latter desires \textit{temporal equivariance} representations. This motivates us to question the suitability of the existing ``TAC features for TAL'' paradigm. Thus, in this section, we delve into the design of unsupervised pre-training customized for TAL, to reach the task alignment goal, \textit{i.e.}, ``TAL features for TAL''. 

For the TAC task, given a video $\bm{v}_i$ from a dataset $\bm{V}=\{\bm{v}_i\}_{i=1}^N$, the goal is to learn a feature encoding function $\mathcal{F}(\bm{v})$ with which the extracted representation is ensured to be insensitive to the temporal transformation $\mathcal{T}$, {\it i.e.} $\forall\bm{v} \in \bm{V}: \mathcal{F}(\mathcal{T}(\bm{v})) = \mathcal{F}(\bm{v})$, as illustrated in Fig.~\ref{fig:in-equi} top part. To achieve this objective, the learning strategy can be basically designed as pushing $\mathcal{F}(\mathcal{T}(\bm{v}))$ and $\mathcal{F}(\bm{v})$ close to each other in the feature space. To be more general, two random transformations $\mathcal{T}$ and $\mathcal{T}'$ are applied to $\bm{v}$ to implement the strategy, and contrastive learning~\cite{oord2018representation} is involved to enforce the consistency:

\begin{equation}
\mathcal{F}(\mathcal{T}(\bm{v})) \mypull \mathcal{F}(\mathcal{T}'(\bm{v})),
\label{equation:CL_push_in}
\end{equation}
in which the identity mapping $\mathcal{T}_0(\bm{v})=\bm{v}$ is also considered.

Under the scenario of TAL, we require $\mathcal{F}$ to be sensitive to the transformation $\mathcal{T}$, {\it i.e.} $\forall\bm{v} \in \bm{V}: \mathcal{F}(\mathcal{T}(\bm{v})) = \mathcal{T}(\mathcal{F}(\bm{v}))$, which can be re-written as $\mathcal{F}(\bm{v}) = \mathcal{G}(\mathcal{F}(\mathcal{T}(\bm{v})))$ and $\mathcal{G} \triangleq \mathcal{T}\inv$ (See Fig.~\ref{fig:in-equi} bottom part). Similar to Eqn.~\ref{equation:CL_push_in}, we apply two random transformations to $\bm{v}$, and therefore have
\begin{equation}
\mathcal{F}(\bm{v}) = \mathcal{G}(\mathcal{F}(\mathcal{T}(\bm{v}))) = \mathcal{G}'(\mathcal{F}(\mathcal{T}'(\bm{v}))).
\label{equation:equivariance}
\end{equation}

Intuitively, contrastive learning can be introduced here to model the temporal equivariance by forcing the features processed by two transformation pairs $(\mathcal{T}, \mathcal{G})$ and $(\mathcal{T}', \mathcal{G}')$ respectively to be analogous to one another:
\begin{equation}
\mathcal{G}(\mathcal{F}(\mathcal{T}(\bm{v}))) \mypull \mathcal{G'}(\mathcal{F}(\mathcal{T}'(\bm{v}))).
\label{equation:CL_push_equi}
\end{equation}

In the following sections, a parameterized temporal transformation $\mathcal{T}$ tailored for TAL tasks is introduced. We delicately design a new self-supervised task called \lmodel (\smodel) with self-generated transformation signals $\mathcal{T}$, and apply the contrastive strategy to learn temporal translation and scale equivariance encoding $\mathcal{F}$.

\subsection{\lmodel}

As illustrated in Fig.~\ref{fig:PAL}, given a large-scale trimmed video dataset (\textit{e.g.}, Kinetics~\cite{carreira2017quo}), we randomly select two temporal regions from one video (viewed as pseudo action regions), and then paste them onto another two videos (viewed as pseudo background) at various scales and locations. With the self-generated temporal locations and scales treated as prior during pre-training, the model is expected to localize the pseudo action regions from the synthesized new videos. Instead of directly predicting the paste locations and scales, we introduce the contrastive strategy to enforce the consistency between the features of two random regions defined by the priors for temporal equivariance representation learning, as illustrated in Eqn.~\ref{equation:CL_push_equi}.

In this pipeline, we first perform transformation $\mathcal{T}$ in the input space for TAL-tailored video generation (Sec.~\ref{sec:trans}). Then we use backbone $\mathcal{F}$ and multiple heads to map the transformed videos into the feature space (Sec.~\ref{sec:encode}). Next, equi-transformation $\mathcal{G}$ is applied to inverse the transformation $\mathcal{T}$ in the feature space (Sec.~\ref{sec:equi-trans}). We finally conduct region contrastive learning for TAL-customized pre-training (Sec.~\ref{sec:train_obj}).

\subsubsection{Input-Level Transformation} \label{sec:trans}

To learn the temporal equivariance encoding function $\mathcal{F}$, we define the transformation $\mathcal{T}$ as a video region sampling and {\it paste} operation. Specifically, given a video $\bm{v}_i$ as pseudo action video as well as a randomly selected video $\bm{v}_n$ as pseudo background, we first sample a random region from action video $\bm{v}_i$, and paste it onto the background video $\bm{v}_n$ to generate a synthesized video $\bm{v}_{i \rightarrow n}$. The input-level transformation $\mathcal{T}$ is then defined as follows:

\begin{equation}
\bm{v}_{i \rightarrow n},s,e = \mathcal{T}(\bm{v}_i,\bm{v}_n),
\label{equation:trans}
\end{equation}

\noindent where $s$ and $e$ represent the start and end \textit{clip}\footnote{Here we perform the temporal transformation in a clip-wise manner to align with the clip-level video encoder.} indices of the pseudo action region in the new video $\bm{v}_{i \rightarrow n}$.

To improve the robustness of the learned representation, we soften the paste operation in the implementation by changing it to a blending one with blending ratio $\beta$, which takes $\beta$ of action region and mix it with $(1-\beta)$ of background region to generate the blended region. $\beta$ is randomized from the range $[0.6, 1]$. Besides, spatial data augmentation is involved to increase the diversity of training data. Following the convention~\cite{qian2021spatiotemporal,han2020self,pan2021videomoco}, we apply random cropping, horizontal flipping, Gaussian blurring and color jittering, and all are temporally consistent. In particular, instead of sampling action regions with fixed stride, we propose a \textit{scale-aware sampling} strategy to add some randomness to the action timescale. Here, we refer to the timescale as how fast an action goes. It stems from the observation that an action video played at different speeds contains almost identical semantics. We simply model the timescale variation by sampling action region frames with different strides. 

Overall, by this sample-and-paste way, our input-level transformation mimics the temporal \textit{location} and \textit{scale} variance in real untrimmed action videos, which also provides a strong supervision signal for TAL-tailored temporal equivariant contrastive learning.
\vspace{-7pt}

\subsubsection{Feature Encoding} \label{sec:encode}

Our feature encoder $\mathcal{F}$ contains a backbone $\mathcal{F}^b$ with non-linear projection head $\mathcal{F}^n$ and a temporal embedding head $\mathcal{F}^t$, namely $\mathcal{F} = \mathcal{F}^b \circ \mathcal{F}^n \circ \mathcal{F}^t$. Formally, given the synthesized video $\bm{v}_{i \rightarrow n}$, the corresponding clip feature sequence $\{\bm{c}_{i \rightarrow n}^{(j)}\}_{j=1}^J$ is obtained by:

\begin{equation}
\begin{gathered}
\{\bm{c}_{i \rightarrow n}^{(j)}\}_{j=1}^J = \mathcal{F}(\bm{v}_{i \rightarrow n}),
\label{equation:feats_seq}
\end{gathered}
\end{equation}

\noindent in which the backbone $\mathcal{F}^b$ is a clip-level encoder, and $\mathcal{F}^t$ is a video-level head for temporal modeling among clips. $J$ is the number of sampled clips. It is noted that applying temporal convolutions ($\mathcal{F}^t$) on chronological clip-level features is crucial in our setting. This enables information aggregation among neighboring clips and therefore the features of pseudo action regions near the boundary can be highly affected by the nearby background. In this way, our \smodel can learn background-insensitive boundary features by maximizing the agreement (Sec.~\ref{sec:train_obj}) between region features of the same video but influenced
by different pseudo backgrounds.

\subsubsection{Feature-Level Equi-Transformation} \label{sec:equi-trans}

Recall that we aim at designing a TAL-tailored pre-training paradigm by extending the contrastive strategy to learn temporal equivariance representations. To this end, we propose to utilize additional free region-wise supervision in the form of inverse temporal transformations. In our case, the changes of pseudo action location in the input composited videos ($\bm{v}_{i \rightarrow n}$) will be reflected in the corresponding ones in their feature sequences ($\{\bm{c}_{i \rightarrow n}^{(j)}\}_{j=1}^J$) obtained by Eqn.~\ref{equation:feats_seq}. 

To echo the input-level transformation $\mathcal{T}$ introduced in Sec.~\ref{sec:trans}, we here define the feature-level equi-transformation $\mathcal{G}$ as an alignment operation. Formally, this feature alignment process is defined as:

\begin{equation}
\{\bm{c}_{i \rightarrow n}^{(j)}\}_{j=s}^{e} = \mathcal{G}(\{\bm{c}_{i \rightarrow n}^{(j)}\}^J_{j=1}, s, e),
\label{equation:equi-trans}
\end{equation}
then the region representation can be obtained by temporally averaging pooling the corresponding sequential clip-level features, \textit{i.e.}, $\bm{r}_{i \rightarrow n}^{(s,e)}={\rm TempAvgPool}(\{\bm{c}_{i \rightarrow n}^{(j)}\}_{j=s}^{e})$.

\subsubsection{Contrastive Training Objective} \label{sec:train_obj}
Following the \textit{transformation} $\mathcal{T}$ and {\it alignment} $\mathcal{G}$ operations introduced above, two pseudo action regions $[s,e]$ and $[s',e']$ from video $\bm{v}_i$ are extracted, and pasted onto two pseudo background videos $\bm{v}_n$ and $\bm{v}_m$ to obtain region representations $\bm{r}_{i \rightarrow n}^{(s,e)}$ and $\bm{r}_{i \rightarrow m}^{(s',e')}$. These two representations are set as the query and positive key pair $(\bm{r}_q, \bm{r}_{k^{+}})$ in contrastive learning, namely $\bm{r}_q = \bm{r}_{i \rightarrow n}^{(s,e)}$ and $\bm{r}_{k^{+}} = \bm{r}_{i \rightarrow m}^{(s',e')}$. The region features from other composited videos are viewed as negatives. 
Given the encoded query $\bm{r}_q$, positive key $\bm{r}_{k^{+}}$ and negatives $\{\bm{r}_{k_i}\}_{i=1}^{K}$, the contrastive learning essentially encourages the query to be similar to the positive sample and dissimilar to the negative ones. Our \smodel is a pretext task and independent of the detailed loss function, so we simply extend the InfoNCE~\cite{oord2018representation} contrastive loss to ensure region consistency in this work:

\begin{equation}
\mathcal{L} = -{\rm log} \frac{{\rm exp}(\bm{r}_q \cdot \bm{r}_{k^{+}}/ \tau)}{{\rm exp}(\bm{r}_q \cdot \bm{r}_{k^{+}}/ \tau)+\sum_{i=1}^K {\rm exp}(\bm{r}_q \cdot \bm{r}_{k_i}/ \tau)},
\label{equation:PAL_push}
\end{equation}

\noindent where $\tau$ is a temperature hyper-parameter and $K$ is the number of negative samples. Equipped with our proposed \smodel, by minimizing the region contrastive loss, the encoding backbone $\mathcal{F}^b$ is encouraged to learn temporal equivariant features, which we believe is beneficial to the TAL tasks.

\section{Experiments}

\subsection{Experimental Settings}

To evaluate our proposed \smodel, we follow the \textit{pre-training} and \textit{transferring} procedures: first pre-train the feature network on a large-scale trimmed dataset without category labels, then transfer the features pre-computed by the frozen backbone to the downstream TAL tasks. 

\subsubsection{Pre-training}

\noindent \textbf{Datasets.} To have an apple-to-apple comparison with other self-supervised video representation learning methods, we use Kinetics~\cite{carreira2017quo} as the initial pre-training dataset, without using any labels. Kinetics is a large-scale trimmed action recognition benchmark. Each video has a single action class and lasts around 10 seconds. The typical version Kinetics-400 (K400) includes $\sim$300k videos with 400 human action classes, and the latest version Kinetics-700 (K700) contains $\sim$650k videos with 700 action classes.

\noindent \textbf{Implementation Details.} We choose I3D~\cite{carreira2017quo}, the commonly used feature encoder in TAL, as the default backbone ($\mathcal{F}^b$) in our experiments. For temporal embedding head ($\mathcal{F}^t$), we employ two-layer temporal convolutions with a kernel size of 3 followed by ReLU activation function. We uniformly sample 8 clips (8 frames per clip) with a resolution of $112\times112$ for each video, and the max clip length of the pseudo action region is limited to 6. The range of the blending ratio $\beta$ is set as $[0.6, 1.0]$. For our scale-aware sampling strategy, the sampling stride of frames within a clip is chosen from $[1, 4]$. Following~\cite{chen2020improved}, we also maintain a memory queue of 16,384 negative samples and use synchronized BN across all layers. We apply L2 norm to the output features from $\mathcal{F}^t$. The temperature $\tau$ is set to 0.07 for all experiments. For optimization, we train our PAL using the Adam algorithm with a weight decay of $10^{-5}$. The initial learning rate is set as $10^{-4}$ and decreases by a factor of 10 when the validation loss saturates. The training takes 200 epochs in total with a batch size of 512 on 64 NVIDIA Tesla V100 GPUs. 

\subsubsection{Transferring to TAL tasks} \label{sec:finetune-TAL}

\noindent \textbf{Target TAL Tasks.} We choose three popular temporal localization tasks to evaluate our \smodel features: Temporal Action Detection (TAD), Action Proposal Generation (APG) and Video Grounding (VG). 

\begin{table*}[!htbp]
\begin{center}
\caption{\textbf{Comparison to state-of-the-art pre-training methods on the target tasks.} We use G-TAD~\cite{xu2020g} and BMN~\cite{lin2019bmn} as the evaluation methods for TAD and APG tasks, respectively. The results are conducted on ActivityNet v1.3 dataset. \bluehl{Rows highlighted in blue use fully-supervised pre-training.} $\dagger$ represents results from~\cite{xu2021low}. * means our implementation. (TR: temporal resolution, SR: spatial resolution) }
\resizebox{1.0\linewidth}{!}{
\def\arraystretch{1.1}
\begin{tabular}{cccccccccccccc}
\toprule
\multirow{2.5}*{\textbf{Method}} & \multirow{2.5}*{\textbf{Modal}} & \multirow{2.5}*{\textbf{Dataset}} & \multirow{2.5}*{\textbf{Backbone}} & \multirow{2.0}*{\textbf{TR$\times$SR$^2$}} & \multirow{2.0}*{\textbf{FLOPs}} &  \multicolumn{4}{c}{\textbf{TAD Task (G-TAD~\cite{xu2020g})}} & \multicolumn{4}{c}{\textbf{APG Task (BMN~\cite{lin2019bmn})}}\\
\cmidrule(lr){7-10} \cmidrule(lr){11-14}
& & & & {\footnotesize \textbf{(per clip)}} & {\footnotesize \textbf{(per clip)}} & mAP@0.5 & @0.75 & @0.95 & AVG & AR@1 & @10 & @100 & AUC \\
\midrule
CoCLR~\cite{han2020self} & V+F & K400 & S3D & 32$\times$128$^2$ & 47.2G & 47.9 & 32.2 & 7.3 & 31.9 & 32.7 & 53.5 & 73.9 & 65.0 \\
XDC~\cite{alwassel2020self} & V+A & IG65M & R(2+1)D-18 & 32$\times$224$^2$ & 325.2G & 48.4 & 32.6 & 7.6 & 32.3 & 33.2 & 54.1 & 74.0 & 65.4  \\
\hdashline
MoCo-v2~\cite{chen2020improved} * & V & K400 & I3D & 8$\times$112$^2$ & 3.6G & 46.6 & 30.7 & 6.3 & 30.3 & 30.8 & 53.5 & 72.4 & 64.0\\
VideoMoCo~\cite{pan2021videomoco} & V & K400 & R(2+1)D-18 & 32$\times$112$^2$ & 81.3G & 47.8 & 32.1 & 7.0 & 31.7 & 31.8 & 53.9 & 72.8 & 65.1 \\
RSPNet~\cite{chen2021rspnet} & V & K400 & R(2+1)D-18 & 16$\times$112$^2$ & 40.6G & 47.1 & 31.2 & 7.1 & 30.9 & 31.5 & 53.3 & 72.2 & 64.1 \\
AoT~\cite{wei2018learning} $\dagger$ & V & K400 & TSM-Res50 & 8$\times$224$^2$ & 33G & 44.1 & 28.9 & 5.9 & 28.8 & - & - & - & -\\
SpeedNet~\cite{benaim2020speednet} $\dagger$ & V & K400 & TSM-Res50 & 8$\times$224$^2$ & 33G & 44.5 & 29.5 & 6.1 & 29.4 & - & - & - & -\\
\textbf{PAL (Ours)} & V & K400 & I3D & 8$\times$112$^2$ & 3.6G & \underline{49.3} & \underline{34.0} & \underline{7.9} & \underline{33.4} & \underline{33.7} & \underline{55.9} & \underline{75.0} & \underline{66.8} \\
\textbf{PAL (Ours)} & V & K700 & I3D & 8$\times$112$^2$ & \textbf{3.6G} & \textbf{50.7} & \textbf{35.5} & \textbf{8.7} & \textbf{34.6} & \textbf{34.2} & \textbf{57.8} & \textbf{76.0} & \textbf{68.1} \\
\midrule
\belowrulesepcolor{aliceblue}
\rowcolor{aliceblue} TAC * & V & K400 & I3D & 8$\times$112$^2$ & 3.6G & 48.5 & 32.9 & 7.2 & 32.5 & 32.3 & 54.6 & 73.5 & 65.6 \\
\rowcolor{aliceblue} BSP~\cite{Xu_2021_ICCV} & V & K400 & TSM-Res50 & 8$\times$224$^2$ & 33G & 50.9 & 35.6 & 8.0 & 34.8 & 33.7 & 57.4 & 75.5 & 67.6\\
\rowcolor{aliceblue} LoFi-E2E~\cite{xu2021low} & V & K400+\textbf{ANet} & TSM-Res18 & 8$\times$224$^2$ & 14.6G & 50.4 & 35.4 & 8.9 & 34.4 & - & - & - & -\\
\rowcolor{aliceblue} TSP~\cite{alwassel2021tsp} & V & K400+\textbf{ANet} & R(2+1)D-34 & 16$\times$112$^2$ & 76.4G & 51.3 & 37.1 & 9.3 & 35.8 & 35.0 & 59.0 & 76.6 & 69.0\\
\aboverulesepcolor{aliceblue}
\bottomrule
\end{tabular}
}
\label{table:TAD_APG_comp}
\end{center}
\vspace{-16pt}
\end{table*}

\noindent \textbf{Datasets.} (1) \textit{ActivityNet v1.3}~\cite{caba2015activitynet} is a popular large-scale benchmark for TAD and APG tasks, including 10,024 training videos, 4,926 validation videos corresponded to 200 action classes. Each video contains 1.65 action instances on average; (2) \textit{Charades-STA}~\cite{gao2017tall} is commonly used for VG task, containing 12,408 and 3,720 text query pairs in training and test set, respectively. The average duration of videos is 30 seconds and the
maximum length of a text query is 10; (3) \textit{THUMOS'14}~\cite{idrees2017thumos} is a standard benchmark for TAD and APG tasks, containing 200 validation videos and 213 test videos of 20 action categories. The video length varies greatly, from less than a second to about 26 minutes. On average, each video contains $\sim$16 action instances. 

\noindent \textbf{Evaluation Metrics.} We follow the standard evaluation protocol. For the TAD task, we report mean Average Precision (mAP) values under different temporal Intersection over Union (tIoU) thresholds. For the APG task, we report the Area Under the Curve (AUC) of the average recall \textit{vs.} average number (AR-AN) of proposals per video. For the VG task, the top-1 recall at three tIoU thresholds and their mean value (mIoU) are reported.

\noindent \textbf{Implementation Details.} To validate the efficacy of our pre-training strategy, we retrain several state-of-the-art TAL methods by only replacing the original features with our \smodel features. We choose those representative works with publicly-available codes. Specifically, we choose G-TAD~\cite{xu2020g} for TAD task, BMN~\cite{lin2019bmn} for APG task, and LGI~\cite{mun2020local} for VG task.

\vspace{-3pt}
\subsection{Main Results}
\vspace{-3pt}

In this section, we compare the performance of our \smodel with other state-of-the-art pre-training approaches on three challenging TAL tasks. For those self-supervised methods designed for TAC tasks, we directly use their released pre-trained models to extract the video features for the downstream TAL task evaluations.

\textbf{Temporal Action Detection (TAD) \& Action Proposal Generation (APG).} In Table~\ref{table:TAD_APG_comp}, we report our TAD and APG results on ActivityNet v1.3 and compare them with state-of-the-art pre-training methods. When pre-trained on K400, our \smodel consistently outperforms other self-supervised methods, which strongly demonstrates the effectiveness of our method. Although these self-supervised pre-training competitors have achieved promising results on TAC tasks, the task discrepancy issue still harms their transferability on TAL tasks, which verifies the necessity of our work. Compared to our baseline MoCo-v2~\cite{chen2020improved}, which focuses on learning temporal invariant features, our proposed temporal equivariant learning scheme is more suitable for TAL, so it yields an improvement of +3.1\% mAP@AVG and +2.8\% AUC gains under the same settings. Notably, when using the same backbone (I3D) and pre-training dataset (K400), our unsupervised \smodel even surpasses the supervised counterpart TAC by gains of +0.9\% on mAP@AVG and +1.2\% on AUC. It suggests that the proper use of data may benefit more than action label annotation information in TAL, which is consistent with our motivation. When pre-trained on a larger dataset K700, our \smodel further improves the performance, showing its potential benefit of leveraging large-scale web videos.
Compared to recent fully-supervised pre-training methods including BSP~\cite{Xu_2021_ICCV}, LoFi-E2E~\cite{xu2021low} and TSP~\cite{alwassel2021tsp}, our first attempt on unsupervised TAL pre-training achieves competitive results. Note that both LoFi-E2E~\cite{xu2021low} and TSP~\cite{alwassel2021tsp} use the downstream dataset ActivityNet (ANet) for feature pre-training which can lead to unfair comparison.

\textbf{Video Grounding (VG).} The VG results on Charades-STA are reported in Table~\ref{table:VG_comp}. Note that the original LGI~\cite{mun2020local} exploits I3D features fine-tuned on the downstream Charades-STA dataset. For fair comparison, we retrain the LGI model using K400 pre-trained I3D features, without changing any hyper-parameters in the original codebase. Clearly, our \smodel achieves the best VG performance under the unsupervised pre-training setting and even surpasses the supervised TAC trained features. Note that BSP~\cite{Xu_2021_ICCV} feature, which is pre-trained in a supervised manner and has much more per-clip FLOPs (33G \textit{vs.} 3.6G), performs better than ours as expected. 

Overall, to our best knowledge, as the first unsupervised pre-training work customized for TAL, \smodel consistently surpasses other unsupervised pre-training methods on three typical TAL tasks, demonstrating the efficacy of our idea.

\begin{table}[t]
\begin{center}
\caption{\textbf{Comparison to state-of-the-art pre-training methods on the VG task.} We use LGI~\cite{mun2020local} as the evaluation method. The results are conducted on Charades-STA dataset. \bluehl{Rows highlighted in blue use fully-supervised pre-training.} * Our implementation. 
}
\resizebox{0.85\linewidth}{!}{
\begin{tabular}{ccccc}
\toprule
\multirow{1.7}*{\textbf{Method}} & \multicolumn{4}{c}{\textbf{VG Task (LGI~\cite{mun2020local})}}\\
\cmidrule(lr){2-5}
\textbf{{\footnotesize (K400 pre-trained)}} & R@0.3 & R@0.5 & R@0.7 & mIoU\\
\midrule
MoCo-v2~\cite{chen2020improved} * & 54.2 & 40.9 & 21.1 & 38.7 \\
VideoMoCo~\cite{pan2021videomoco} & 59.1 & 44.5 & 23.4 & 42.3 \\
RSPNet~\cite{chen2021rspnet} & 55.8 & 41.5 & 21.4 & 39.6 \\
\textbf{\smodel (Ours)} & \textbf{63.7} & \textbf{50.0} & \textbf{27.2} & \textbf{46.8} \\
\midrule
\belowrulesepcolor{aliceblue}
\rowcolor{aliceblue} TAC * & 61.6 & 46.8 & 24.6 & 44.3 \\
\rowcolor{aliceblue} BSP~\cite{Xu_2021_ICCV} & 68.8 & 53.6 & 29.3 & 50.6 \\
\aboverulesepcolor{aliceblue}
\bottomrule
\end{tabular}
}
\label{table:VG_comp}
\end{center}
\vspace{-16pt}
\end{table}

\subsection{Ablation Study}

In this section, we conduct ablation experiments to fully understand the concept of our \smodel. For ease of experimentation, all the ablation studies are conducted with 100 training epochs on K400 and evaluated on the TAD task.

\textbf{Effectiveness of the key \smodel components.} In Table~\ref{table:abla_components}, we examine how each design in \smodel affects the overall performance. We consider three key components in \smodel: (1) dense sampling strategy to select multiple clips as region sample; (2) scale-aware sampling strategy to sample pseudo action regions with different strides; (3) paste operation to paste the selected regions onto the background videos. We start with the basic setting that does not involve any of the above designs, where only one clip is randomly sampled from each video and clip-level contrastive learning is performed. Then we introduce the dense sampling strategy to contrast region-level embeddings, which brings +0.5\% improvement due to more temporal clues being included. Next, a scale-aware sampling strategy is applied along with dense sampling, leading to an overall +1.1\% gain. This verifies that adding some randomness to the  temporal scale facilitates representation learning. The biggest improvement is achieved after introducing paste operation. 
We infer that this is because attracting the region features of the same video but influenced by different backgrounds yields more background-insensitive features, which benefits localization tasks.  
Finally, pasting pseudo action regions onto different background videos further contributes a reasonable gain of +0.7\% and the final improvement reaches +3.2\% compared with baseline. In summary, by adding these key components step by step, the performance consistently boosts, verifying the effectiveness of our \smodel. 

\begin{table}[t]
\begin{center}
\caption{\textbf{Contribution of each design in \smodel on TAL tasks.} Dense sampling strategy (dense), scale-aware sampling strategy (scale) and paste operation (paste) are involved step by step. All these designs contribute to the overall performance. } 
\resizebox{0.85\linewidth}{!}{
\begin{tabular}{ccccc}
\toprule
\multirow{2.5}*{\textbf{Exp.}} & \multicolumn{3}{c}{\textbf{Setting}} & \textbf{TAD Task}\\
\cmidrule(lr){2-4}
& dense & scale & paste & mAP@AVG\\
\midrule
\#0 (\textit{Baseline}) & \xmark & \xmark & \xmark & 29.6\\
\#1 & \checkmark & \xmark & \xmark & 30.1 {\footnotesize \textbf{\textcolor{purple}{(+0.5)}}}\\
\#2 & \checkmark & \checkmark & \xmark & 30.7 {\footnotesize \textbf{\textcolor{purple}{(+1.1)}}}\\
\#3 & \checkmark & \checkmark & same bkg. & 32.1 {\footnotesize \textbf{\textcolor{purple}{(+2.5)}}}\\
\#4 (PAL) & \checkmark & \checkmark & diff bkg. & \textbf{32.8} {\footnotesize \textbf{\textcolor{purple}{(+3.2)}}}\\
\bottomrule
\end{tabular}
}
\label{table:abla_components}
\end{center}
\vspace{-10pt}
\end{table}

\textbf{Number of the temporal embedding head layers.} In Table~\ref{table:t_embed}, we experiment with the different numbers of the temporal embedding head layers. When the layer number is 0, the input clips are processed independently without temporal fusion, and therefore the surrounding background clips have no substantial effect on the action regions. It's obvious that attaching a single temporal convolution layer atop the backbone significantly boosts the performance (+1.3\%). This verifies our hypothesis that introducing background semantics can help promote the  localization power required by TAL. Since using two temporal embedding layers yields the best performance, we choose this setting as our default.

\begin{figure}[!t]
\noindent
\begin{minipage}{\linewidth}
	\begin{minipage}{0.45\linewidth}
	\captionof{table}{\textbf{Ablation study on temporal embedding head.}}
	\vspace{-16pt}
		\begin{center}
			\resizebox{1.1\linewidth}{!}{
			\begin{tabular}{ccc}
				\toprule
				\textbf{Num.} & \textbf{Recep.} & \textbf{mAP@AVG}\\
				\midrule
				0 & 1 & 30.8 \\
				1 & 3 & 32.1 {\footnotesize \textbf{\textcolor{purple}{(+1.3)}}} \\
				2 & 5 & \textbf{32.8} {\footnotesize \textbf{\textcolor{purple}{(+2.0)}}} \\
				3 & 7 & 32.5 {\footnotesize \textbf{\textcolor{purple}{(+1.7)}}} \\
				\bottomrule
				\end{tabular}
				}
		\end{center}
		\label{table:t_embed}
	\end{minipage}
	\hspace{0.5cm}
	\begin{minipage}{0.45\linewidth}
	\captionof{table}{\textbf{Ablation study on paste ratios.}}
	\vspace{-16pt}
		\begin{center}
			\resizebox{0.8\linewidth}{!}{
			\begin{tabular}{ccc}
			\toprule
			\textbf{$\beta$} & \textbf{mAP@AVG}\\
			\midrule
			$1$ & 32.2 \\
			$[0, 0.4]$ & 30.7 {\footnotesize \textbf{\textcolor{gray}{(-1.5)}}}\\
			$[0.4, 0.6]$ & 31.5 {\footnotesize \textbf{\textcolor{gray}{(-0.7)}}}\\
			$[0.6, 1.0]$ & \textbf{32.8} {\footnotesize \textbf{\textcolor{purple}{(+0.6)}}}\\
			$[0, 1.0]$ & 31.3 {\footnotesize \textbf{\textcolor{gray}{(-0.9)}}}\\
			\bottomrule
			\end{tabular}
			}
		\end{center}
		\label{table:beta}
	\end{minipage}
\end{minipage}
\vspace{-10pt}
\end{figure}

\textbf{Hard paste \textit{vs.} Soft paste.} In \smodel, $\beta$ controls the paste ratio of pseudo action regions onto the background videos. We evaluate different $\beta$ from 0 to 1. In particular, $\beta=1$ means the ``hard'' paste and $\beta<1$ is the ``soft'' paste. For the soft way, we test several representative intervals: $[0, 0.4]$, $[0.4, 0.6]$, $[0.6, 1.0]$ and $[0, 1.0]$, which indicates four cases, {\it i.e.}, background-dominated, half-and-half, action-dominated and purely random, respectively. As shown in Table~\ref{table:beta}, the $\beta \in [0.6, 1.0]$ setting outperforms hard paste and achieves the best result, partially because the action-dominated soft paste serves as an effective data augmentation strategy. So we use this setting by default. 

\begin{table}[t]
\begin{center}
\caption{\textbf{TAD results on small-scale THUMOS'14 dataset.} 
}
\vspace{-0.15cm}
\resizebox{0.95\linewidth}{!}{
\begin{tabular}{cccccc}
\toprule
\multirow{1.7}*{\textbf{Method}} & \multicolumn{5}{c}{\textbf{TAD Task (G-TAD~\cite{mun2020local})}}\\
\cmidrule(lr){2-6}
\textbf{{\footnotesize (K400, 100 epochs)}} & mAP@0.3 & @0.4 & @0.5 & @0.6 & @0.7\\
\midrule
\belowrulesepcolor{aliceblue}
\rowcolor{aliceblue} TAC * & 44.6 & 37.3 & 29.5 & 18.8 & 9.5 \\
MoCo-v2~\cite{chen2020improved} * & 41.5 & 34.1 & 25.8 & 17.3 & 7.9\\
\midrule
\textbf{\smodel (Ours)} & \textbf{46.8} & \textbf{40.3} & \textbf{30.8} & \textbf{19.3} & \textbf{10.9} \\
\bottomrule
\end{tabular}
}
\label{table:TH14_comp}
\end{center}
\vspace{-10pt}
\end{table}

\begin{table}[t]
\begin{center}
\caption{\textbf{Comparison on TAC task.} 
}
\vspace{-0.15cm}
\resizebox{0.85\linewidth}{!}{
\begin{tabular}{cccc}
\toprule
\multirow{1.7}*{\textbf{Method}} & \multirow{2.5}*{\textbf{Backbone}} & \multicolumn{2}{c}{\textbf{TAC Task {\footnotesize (Top-1 Acc.)}}}\\
\cmidrule(lr){3-4}
\textbf{{\footnotesize (K400, pre-trained)}} & & \textbf{UCF101} & \textbf{HMDB51}\\
\midrule
MoCo-v2~\cite{chen2020improved} * & 3D-Res50 & 82.0 & 49.4\\
AoT~\cite{wei2018learning} & T-CAM & 79.4 & - \\
SpeedNet~\cite{benaim2020speednet} & S3D-G &81.1 & 48.8 \\
VTHCL~\cite{yang2020video} & 3D-Res50 & 82.1 & 49.2 \\
VideoMoCo~\cite{pan2021videomoco} & R(2+1)D-18 & 78.7 & 49.2 \\
CoCLR~\cite{han2020self} & S3D & \textbf{87.9} & \textbf{54.6} \\
\textbf{\smodel (Ours)} & 3D-Res50 & 84.7 & 52.5\\
\bottomrule
\end{tabular}
}
\label{table:cls_comp}
\end{center}
\vspace{-15pt}
\end{table}

\textbf{Evaluation on THUMOS'14.} 
THUMOS'14 is a relatively small-scale dataset compared with ActivityNet v1.3 (\textit{cf.} Sec.~\ref{sec:finetune-TAL}).
We list the experimental results in Table~\ref{table:TH14_comp}. Compared with the relative performance gain on ActivityNet v1.3, our improvement on THUMOS'14 is more prominent, which confirms the generalization ability of \smodel under the small-scale data condition. 

\textbf{Evaluation on TAC task.} We investigate the transferring ability of our PAL on TAC downstream task. Following the common practice, all layers are fine-tuned end-to-end. The results are evaluated on UCF101~\cite{soomro2012ucf101} and HMDB51~\cite{kuehne2011hmdb} datasets. Although our \smodel feature is designed for TAL tasks, we observe in Table~\ref{table:cls_comp} that it still achieves competitive performance on TAC tasks. In detail, our \smodel outperforms baseline MoCo-v2 by +2.7\% \& +3.1\% at top-1 accuracy on UCF101 and HMDB51 respectively. Notably, it even exceeds the recently proposed VTHCL~\cite{yang2020video} with the same backbone. 

\subsection{Feature Visualization}

Recall that \smodel is proposed to guide the network in learning temporal translation and scale equivariance ability. To confirm this, we apply temporal transformations on the action instances in real-world videos and investigate whether these changes will be reflected in feature space accordingly. Specifically, given a video from ActivityNet v1.3, we first crop the action instance based on the temporal annotations, then re-sample the action instance with different temporal strides and insert them back into random temporal locations. Here, we consider two temporal transformations: (1) \textit{2$\times$ down-sampling} the action instance and moving \textit{backward} along the time axis; and (2) \textit{2$\times$ up-sampling} the action instance and moving \textit{forward} along the time axis. Next, MoCo-v2~\cite{chen2020improved} (baseline), VideoMoCo~\cite{pan2021videomoco} and our \smodel encoders are applied to extract features for the original video and the two transformed videos. 

\begin{figure}[!t]
\begin{center}
\begin{subfigure}[b]{0.8\linewidth}
         \centering
\includegraphics[width=\linewidth]{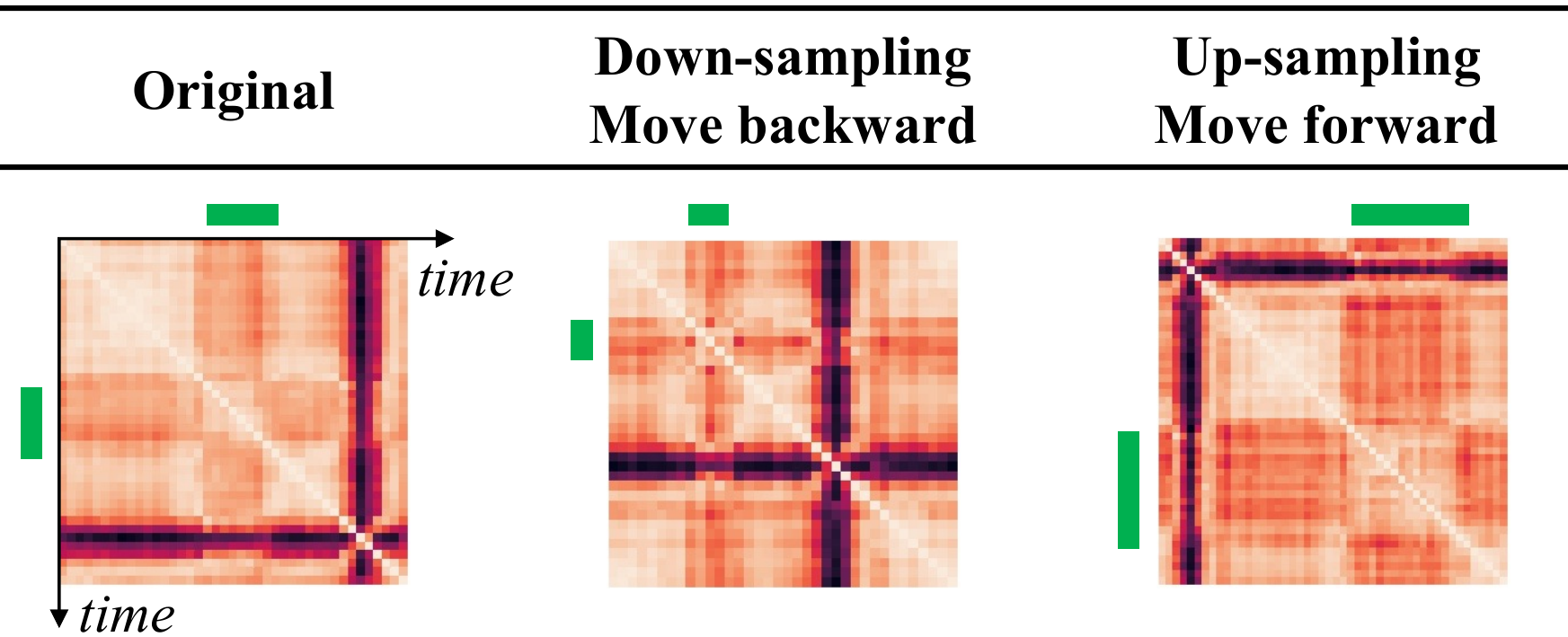}
\caption{MoCo-v2~\cite{chen2020improved} (baseline)}
\end{subfigure}
\begin{subfigure}[b]{0.8\linewidth}
         \centering
\includegraphics[width=\linewidth]{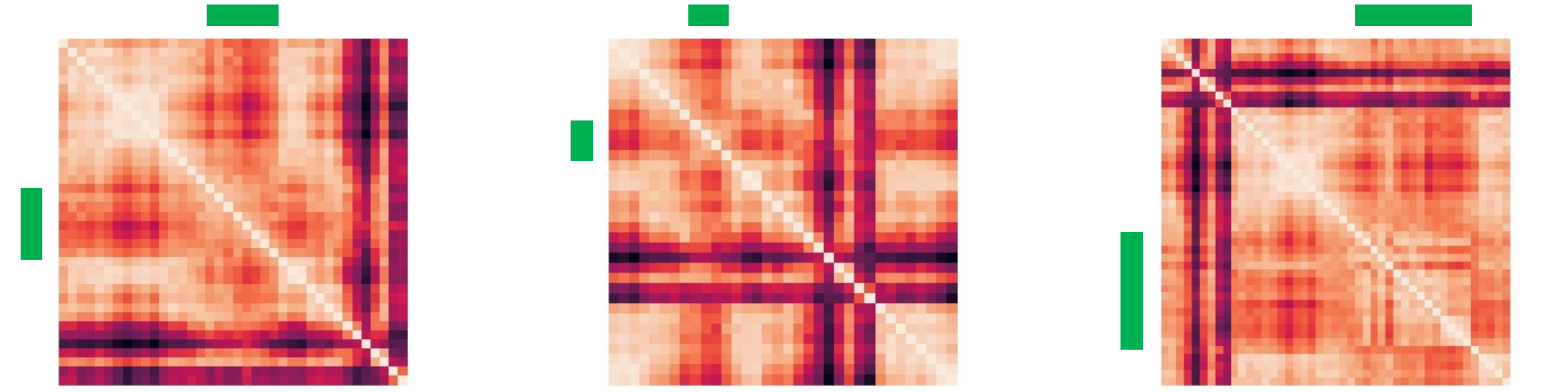}
\caption{VideoMoCo~\cite{pan2021videomoco}}
\end{subfigure}
\begin{subfigure}[b]{0.8\linewidth}
         \centering
\includegraphics[width=\linewidth]{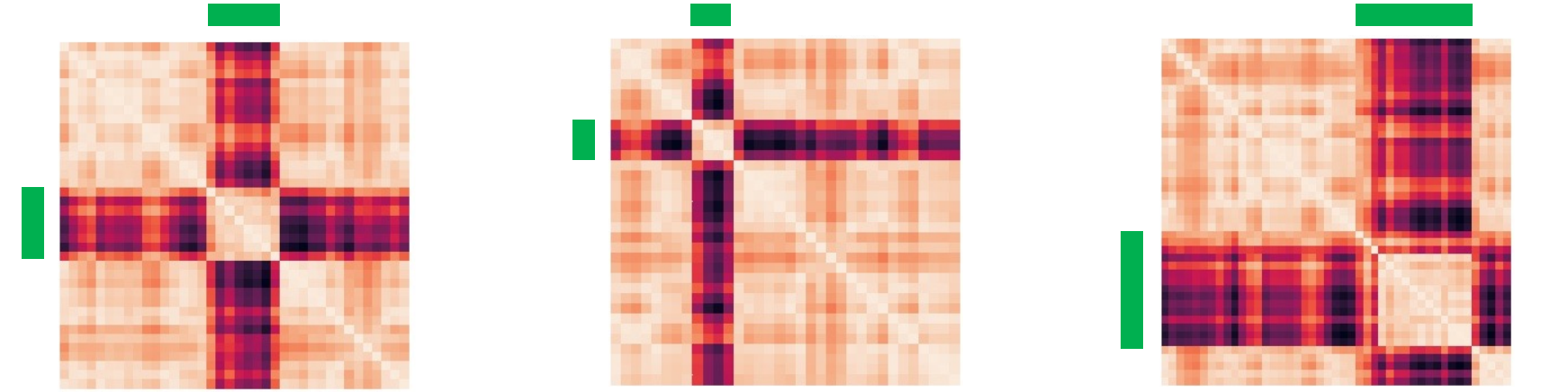}
\caption{\textbf{PAL (Ours)}}
\end{subfigure}
\begin{subfigure}[b]{0.8\linewidth}
         \centering
\includegraphics[width=\linewidth]{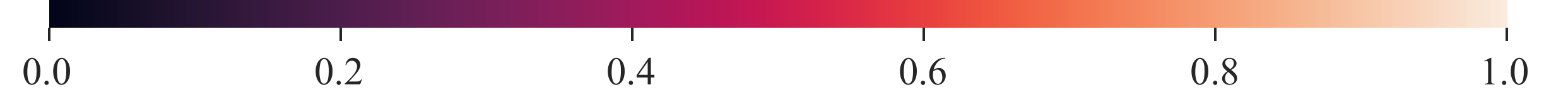}
\end{subfigure}
\end{center}
\vspace{-0.5cm}
 \caption{\textbf{Feature similarity visualization under different temporal transformations (2$^{\bm nd}$ \& 3$^{\bm rd}$ columns) of ground-truth action instance.} The green bars represent the temporal extent of ground truth actions. Brighter color means higher similarity. }
\label{fig:vis_sim}
\end{figure}

We visualize the cosine similarity between each clip features pair within the same video in Fig.~\ref{fig:vis_sim}. We also plot the ground-truth annotations (green bars) to indicate the action clips. As can be seen, MoCo-v2 and VideoMoCo learn time-invariant features that are insensitive to the temporal transformations. There is a high similarity between the pseudo action region and the background, while the salient area of \smodel features changes accordingly.
This confirms that our method successfully learns the time-equivariant characteristic, which is naturally more beneficial for TAL tasks. Besides, our introduced time-equivariant learning scheme can not only better separate the action and background clips, but also enable sharper contrast between action and its surrounding background clips. In this way, the clip features become more informative and boundary-aware, which facilitates localization. More visualization results can be found in our supplementary materials. 

\vspace{-3pt}
\section{Discussion and Conclusion}
\vspace{-3pt}

This paper presents a new pretext task called \lmodel (\smodel), which is delicately designed to pre-train representations in an unsupervised manner for TAL tasks (UP-TAL). Motivated by the essential discrepancy between TAC and TAL, we also introduce a temporal scale and location equivariance learning scheme to facilitate better task alignment for the downstream transferring process. On a variety of downstream TAL tasks including temporal action detection, action proposal generation and video grounding, we demonstrate the effectiveness of our proposed method, which consistently surpasses its TAC counterpart and other unsupervised pre-training methods. 

{\footnotesize \noindent \textbf{Acknowledgements.} This paper was partially supported by the National Natural Science Foundation of China (NSFC) 62176008. Special acknowledgements are given to Aoto-PKUSZ Joint Lab for its support.}

\clearpage

%%%%%%%%% REFERENCES
{\small
\bibliographystyle{ieee_fullname}

}

\end{document}